# GSTurb: Gaussian Splatting for Atmospheric Turbulence Mitigation

Hanliang Du, Zhangji Lu, Zewei Cai, Qijian Tang, Qifeng Yu, Xiaoli Liu

*Abstract*—Atmospheric turbulence causes significant image degradation due to pixel displacement (tilt) and blur, particularly in long-range imaging applications. In this paper, we propose a novel framework for atmospheric turbulence mitigation, GSTurb, which integrates optical flow-guided tilt correction and Gaussian splatting for modeling non-isoplanatic blur. The framework employs Gaussian parameters to represent tilt and blur, and optimizes them across multiple frames to enhance restoration. Experimental results on the ATSyn-static dataset demonstrate the effectiveness of our method, achieving a peak PSNR of 27.67 dB and SSIM of 0.8735. Compared to the state-of-the-art method, GSTurb improves PSNR by 1.3 dB (a 4.5% increase) and SSIM by 0.048 (a 5.8% increase). Additionally, on real datasets, including the TSRWGAN Real-World and CLEAR datasets, GSTurb outperforms existing methods, showing significant improvements in both qualitative and quantitative performance. These results highlight that combining optical flow-guided tilt correction with Gaussian splatting effectively enhances image restoration under both synthetic and real-world turbulence conditions. The code for this method will be available at https://github.com/Duhl-Liamz/3DGS_turbulence/tree/main.

## I. INTRODUCTION

ATMOSPHERIC turbulence is a significant degradation factor in long-range imaging systems, caused by random spatial and temporal refractive index fluctuations in the atmosphere during wavefront propagation [1]. These disturbances result in two primary effects: image pixel displacement (tilt) and image blur, which substantially degrade image quality and hinder the accurate detection, recognition, and measurement of distant targets. As a result, mitigating atmospheric turbulence is essential for applications in free-space optical communication [2], astronomical imaging [3], and remote sensing [4].

Atmospheric turbulence mitigation aims to recover clear images from degraded ones, a typical ill-posed problem. The complexity of this problem is further increased due to its high randomness. To address this challenge, it is essential to estimate key factors like tilt and blur in the mitigation process.

Traditional mitigation methods for tilt correction mostly rely on feature registration or conventional optical flow computation [5]. These methods use a reference frame image as the baseline, which is assumed to be free from tilt, enabling tilt correction in the degraded images.

Image blur caused by atmospheric turbulence can be considered as having a spatially invariant blur kernel in small fields of view (FOV), i.e., isoplanatic imaging. However, in large fields of view, it becomes non-isoplanatic, with the blur kernel varying spatially [6]. Traditional deblur methods typically assume isoplanatic imaging conditions and employ blind deconvolution techniques [7], [8], [9], [10]. For non-isoplanatic imaging in large FOV, more complex models and lucky frames are required [11], [12].

With the development of atmospheric turbulence image simulation and synthesis algorithms [13], [14], data-driven deep learning mitigation algorithms have gradually become mainstream [15], [16], [17], [18]. Due to the ill-posed problem and randomness of the atmospheric turbulence mitigation problem, inputting multi-frame long sequences is crucial for effective restoration.

Existing methods suffer from several limitations. Traditional tilt correction techniques, while useful, rely heavily on the assumption that a reference frame image is tilt-free, which can lead to biases when the reference itself contains tilt. In deblurring, traditional methods assume isoplanatic conditions that are not suitable for large fields of view, requiring complex models and lucky frames, which are both unstable and computationally expensive. In deep learning-based methods, although they have shown promise, model capacity limitations prevent them from processing large image batches effectively, leading to bottlenecks in restoration quality. Additionally, these methods often fail to sufficiently incorporate the physical processes underlying atmospheric turbulence, which limits their ability to generalize and accurately model turbulence degradation.

In recent years, Gaussian Splatting technology has

This work is supported by National Natural Science Foundation of China (62275173, 62371311,62505189); Fundamental Research Project of Shenzhen Municipality (JCYJ20220531101204010); Scientific Instrument Development Project of Shenzhen University (2023YQ009); Shenzhen University Research Team Cultivation Project (2023JCT003). *(Corresponding author: Zhangji Lu, Xiaoli Liu. Co-first author: Zhangji Lu).*
Hanliang Du is with the Faculty of Science and Technology, University of Macau, Macau 999078, China (e-mail: yc27958@um.edu.mo).
Zhangji Lu is with the College of Physics and Optoelectronic Engineering, Shenzhen University, Shenzhen 518000, China (e-mail: zjlu@szu.edu.cn)
Zewei Cai is with the College of Physics and Optoelectronic Engineering, Shenzhen University, Shenzhen 518000, China (e-mail: caizewei@szu.edu.cn)
Qijian Tang is with the College of Physics and Optoelectronic Engineering, Shenzhen University,Shenzhen 518000, China (e-mail: ytqjtang@163.com)
Qifeng Yu is with College of Physics and Optoelectronic Engineering, Shenzhen University, Shenzhen, 518060, China, State Key Laboratory of Radio Frequency Heterogeneous Integration (Shenzhen University), Shenzhen, 518000, China, Shenzhen Key Laboratory of Intelligent Optical Measurement and Detection, Shenzhen University, Shenzhen, 518000, China, Hunan Provincial Key Laboratory of Image Measurement and Vision Navigation,Changsha,410073, China, and also with College of Aerospace Science and Engineering, National University of Defense Technology, Changsha, 410073, China (e-mail: yuqifengszu@163.com)
Xiaoli Liu is with the College of Physics and Optoelectronic Engineering, Shenzhen University, Shenzhen 518000, China (e-mail: lxl@szu.edu.cn)

demonstrated unique advantages in scene reconstruction [10], which may provide new solutions for atmospheric turbulence mitigation. Gaussian Splatting simulates the physical process of projecting an image from a real object, and by embedding atmospheric turbulence degradation into this process, a complete atmospheric turbulence imaging framework can be formed. Gaussian splatting uses a set of Gaussians to fit the real object, with each Gaussian parameter being independent of the others. This property greatly facilitates modeling non-isoplanatic imaging. The Gaussian parameters include position, scale, rotation, transparency, and color, and these parameters allow both the tilt and blur in atmospheric turbulence degradation to be incorporated into the Gaussian splatting optimization process. This simplifies turbulence modeling and provides a straightforward and effective framework for mitigation.

Although 3D Gaussian Splatting (3DGS) has shown promising results for visible light scenes, several challenges remain when applying it directly to atmospheric turbulence mitigation:

- A mapping model needs to be established between the atmospheric turbulence degradation process and the Gaussian splatting parameters. By optimizing the Gaussian splatting parameters, the degradation process can be recovered. Specifically, modeling under non-isoplanatic imaging conditions requires an effective representation of the high randomness of the turbulence degradation process.
- Once the mapping model is established, it is necessary to estimate the tilt parameters and the random blur kernel parameters during the degradation process.
- The physical process of atmospheric turbulence imaging must be established, embedding the degradation represented by the Gaussian Splatting parameters into the entire physical process to ensure that the recovery process aligns with the physical laws of atmospheric turbulence.

To address the challenges described above, this paper introduces a novel algorithmic framework for atmospheric turbulence mitigation, GSTurb, based on Gaussian Splatting. The primary contributions of this work are as follows:

- We propose an atmospheric turbulence mitigation framework based on Gaussian Splatting (GSTurb). To the best of our knowledge, it is the first time that Gaussian Splatting is introduced to realize atmospheric turbulence mitigation. We establish a mapping model between Gaussian Splatting parameters and the atmospheric turbulence degradation process, representing tilt and blur using Gaussian Splatting parameters. This enables the mitigation process to be unified into a simple and effective framework. Due to the efficient representation and computation of Gaussian Splatting, this framework breaks through the input image quantity limitations of existing deep learning models, significantly increasing the number of input images and further improving the quality of image restoration.
- We innovatively propose a simplified method that applies the Recurrent All-Pairs Field Transforms (RAFT) model [11] to the tilt correction problem in atmospheric turbulence degradation. RAFT is a deep learning architecture specifically designed for optical flow estimation, significantly enhancing the accuracy and efficiency of optical flow estimation by combining cyclic update mechanisms and multi-scale correlation volumes. In this study, we combine the optical flow estimation results with the statistical priors of atmospheric turbulence, enabling a simple and efficient tilt correction method. This approach quickly quantifies the tilt degree by estimating the optical flow field between images and utilizes the statistical characteristics of atmospheric turbulence for precise tilt correction. Compared to traditional methods, this strategy, which combines optical flow estimation with statistical priors, not only simplifies the computational process but also significantly improves accuracy, providing a more efficient solution for atmospheric turbulence image restoration.
- We propose a blur kernel estimation network (BKENet) based on ResNeXt [12] to address the complex and randomly varying blur kernels caused by atmospheric turbulence. Using Principal Component Analysis (PCA), we decompose the blur kernels into one principal component basis function and 100 subcomponent basis functions [8]. Our blur kernel estimation network estimates the weights of the subcomponent basis functions and computes the weighted blur kernel. This approach significantly reduces the complexity of estimating random turbulence blur kernels, and when combined with the previously discussed non-isoplanatic imaging modeling, it better represents the spatial randomness of the turbulence blur kernels. By embedding this model into the GSTurb framework, the entire optimization process aligns more closely with the physical processes of atmospheric turbulence imaging, making the optimization results no longer limited to the dataset and enhancing the generalization ability of the model.

II. RELATED WORK

Atmospheric turbulence mitigation can be categorized into two main approaches: traditional methods and learning-based methods. In this section, we provide a brief overview of both categories.

*A. Traditional Turbulence Mitigation Methods*

Atmospheric turbulence mitigation can be decoupled into the correction of tilt and the restoration of image blur [14]. The basic steps of traditional turbulence mitigation methods are as follows: First, multi-frame turbulence images are fused in sequence, and pixel registration is performed to correct tilt of the pixels. Then, under the assumption of isoplanatic imaging, where the image blur is spatially invariant, blind deconvolution methods can be applied to restore the turbulence-induced blur.

For tilt correction in static scenes, traditional methods typically use frame averaging. That is, assuming the tilt induced by turbulence follow a zero-mean distribution over time [15], [16], [17], a multi-frame averaged image is selected as the reference image, which is considered free from tilt. Other frame images are then registered to this reference image as the baseline. Registration methods such as B-splines or optical flow can be used [14], [18]. Due to the randomness of atmospheric turbulence, there are areas in the degraded image that are unaffected by turbulence, approaching the diffraction limit, known as the "lucky frame effect" [19]. Vorontsov et al. select the clearest frame from the video as the lucky frame [20]. However, existing traditional methods for lucky frame selection require complex modeling and computation, and in most cases, it is difficult to find a perfectly suitable lucky frame when the data size is small. On the other hand, using methods like optical flow for registration requires extensive computation, and the reference image is often not perfectly tilt-free.

After tilt correction of the image, the blur can then be restored. In traditional methods, atmospheric turbulence imaging is generally treated as isoplanatic imaging, where the blur kernel is spatially invariant [14]. Therefore, blind deconvolution can be applied to deblur the image. However, this method is only suitable for small-field imaging. For large-field imaging, where the blur kernels across different regions of the image are inconsistent, the isoplanatic assumption fails [21], [22]. Some turbulence mitigation methods still use existing blind deconvolution algorithms, but these methods are nearly unrelated to the statistical properties of turbulence. Mao et al. [7] propose a point spread function (PSF) prior based on turbulence physical constraints, which achieved good results in atmospheric turbulence mitigation under long-range non-isoplanatic conditions. In practice, traditional numerical optimization methods are still commonly used for PSF estimation, but they may suffer from modeling inaccuracies and the ill-posed nature of turbulence degradation. These factors can lead to errors in the estimated PSF and may in turn affect the final restoration quality.

*B. Deep Learning-based Mitigation Methods*

With the development of atmospheric turbulence simulation algorithms [8], [23], significant progress has been made in addressing the issue of lacking ground truth during the training of deep learning models for atmospheric turbulence mitigation. Methods such as [24], [25] have employed generic deblurring networks to implement end-to-end single-frame turbulence-degraded image recovery. Additionally, some approaches [26], [27] have split the recovery process into two parts: de-tilting and de-blurring, using simulated datasets to separately train the de-tilting network and the de-blurring network. Zhang et al. [28] propose the Deep Atmospheric Turbulence Mitigation Network (DATUM), which divides the recovery process into feature registration, temporal fusion, and post-processing, using a physics-based synthetic dataset to improve the model's generalization ability. End-to-end methods have shown promising results, but their model capacity can still pose challenges when handling large image batches, which may limit restoration quality. In addition, these approaches have not yet fully incorporated the physical processes of atmospheric turbulence imaging, which may reduce their generalization ability beyond the training dataset and make it difficult to fully exploit the underlying physical mechanisms of turbulence degradation.

*C. Gaussian Splatting*

3D Gaussian splatting (3DGS) [10] is initially used for 3D scene reconstruction. It models 3D point clouds as Gaussian sphere parameters, optimizing these parameters through differentiable rendering to achieve high-quality 3D scene model reconstruction. Lee et al. [29] associate the blur effect with the scale and opacity of the Gaussian spheres, successfully reconstructing a clear 3D model even in the presence of blurry input images. Zhao et al. [30] combine Bundle Adjusted and 3DGS, removing motion blur from the image during model reconstruction. However, existing 3DGS deblurring methods mainly address defocus blur or motion blur, and research on atmospheric turbulence blur remains sparse. Additionally, the restoration of spatially varying point spread functions (PSF) still requires further research. The method proposed in this paper is the first atmospheric turbulence mitigation framework based on Gaussian Splatting.

III. METHODOLOGY

In this section, we propose a novel framework for atmospheric turbulence mitigation, where the core focus is on the optimization of Gaussian Splatting (GS) parameters. The Tilt Correction Module and Blur Kernel Estimation Module serve as supplementary steps to provide reliable initializations for the GS parameter optimization process.

*A. Overview*

The Optimization Module (Fig. 1(a)) is the core of the framework. It iteratively optimizes the GS parameters, including position, opacity, and scale, using a cyclic consistency loss function. This module refines the Gaussian representation of the turbulence-degraded image, resulting in a sharp, restored image.

The Tilt Correction Module (Fig. 1(b)) begins by using the RAFT model to estimate the optical flow between the reference image and the degraded images. These optical flow fields are averaged based on the zero-mean prior of turbulence to correct the tilt in the reference image. The corrected tilt-free image serves as the input for the next stage of deblurring.

In the Blur Kernel Estimation Module (Fig. 1(c)), BKENet is used to predict the weights of blur kernel basis functions. These weights are linearly combined to form a spatially varying blur kernel, which is then convolved with the tilt-free image to produce a predicted blurred image. The cyclic consistency loss is minimized to further optimize the GS parameters.

By combining the Optimization Module, Tilt Correction Module, and Deblur Module, the framework efficiently restores images degraded by atmospheric turbulence. The GS parameter optimization at the core ensures high-quality restoration by refining both tilt and blur estimates.

*B. GS Turbulence Degradation Modeling*

Based on the GS framework [10], we generate Gaussian distributions, each defined by a distinct set of parameters: three-dimensional position $x$, opacity $\alpha$, a covariance matrix derived from a quaternion $r$ and a scaling factor $s$, and a RGB color factor $c$. However, as illustrated in Fig.1(a), the process of atmospheric turbulence mitigation is primarily related to the parameters $x$, $\alpha$, $r$, $s$. The detailed mathematical formulation is presented in the following section.

In the differentiable rasterization method proposed by GS, each Gaussians is defined by its covariance matrix $\Sigma(r, s)$ and its position $x$ in the world coordinate system, and can be formally expressed as:

$$\mathbf{G}(x,r,s) = e^{-\frac{1}{2}x^T \Sigma^{-1}(r,s)x} \quad (1)$$

The covariance matrix must satisfy the positive semi-definite constraint, which is difficult to ensure during optimization. Therefore, the covariance matrix is decomposed into two learnable parameters: the quaternion $r$ for rotation and the vector $s$ for scaling. This decomposition is similar to ellipsoid parameterization, effectively bypassing the positive semi-

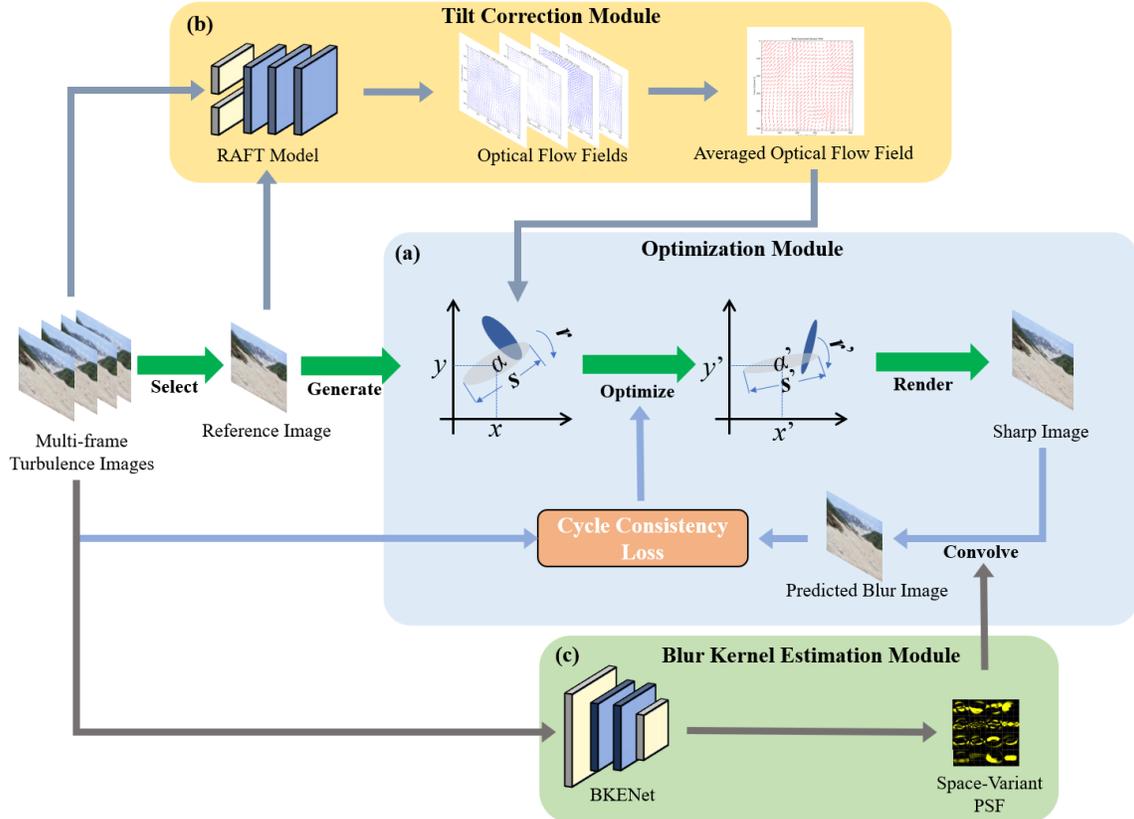

Fig. 1: Gaussian Splatting Atmospheric Turbulence Mitigation Work Flow. Fig. 1(a) illustrates the Optimization Module. Fig. 1(b) illustrates the Tilt Correction Module. Fig. 1(c) illustrates the Blur Kernel Estimation Module.

-definite constraint. Specifically, $r$ and $s$ are transformed into a rotation matrix and a scaling matrix, respectively, and the covariance matrix is constructed as follows:

$$\Sigma(r,s) = \mathbf{R}(r)\mathbf{S}(s)\mathbf{S}(s)^T\mathbf{R}(r)^T \quad (2)$$

where, $\mathbf{R}(r)$ is the rotation matrix corresponding to the quaternion $r$, and $\mathbf{S}(s)$ is the scaling matrix corresponding to the scaling vector $s$.

These Gaussians are projected onto 2D space through a projection transformation. The resulting 2D covariance matrix $\Sigma'(r, s)$ after projection is computed as follows:

$$\Sigma'(r,s) = \mathbf{J}\mathbf{W}\Sigma(r,s)\mathbf{W}^T\mathbf{J}^T \quad (3)$$

Fig. 1: Gaussian Splatting Atmospheric Turbulence Mitigation Work Flow. Fig. 1(a) illustrates the Optimization Module. Fig. 1(b) illustrates the Tilt Correction Module. Fig. 1(c) illustrates the Blur Kernel Estimation Module.

where, $\mathbf{J}$ is the Jacobian matrix of the affine approximation of the projection transformation, and $\mathbf{W}$ is the transformation matrix from the world coordinate system to the camera coordinate system.

The color of each pixel is calculated by accumulating the contributions from $N$ ordered projected 2D Gaussian distributions that overlay on that pixel:

$$\begin{cases} \mathbf{C} = \sum_{i \in N} \mathbf{T}_i c_i \alpha_i \\ \mathbf{T}_i = \prod_{j=1}^{i-1}(1-\alpha_j) \end{cases} \quad (4)$$

where, $c_i$ represents the color of each point, $\mathbf{T}_i$ is the projection rate, and the opacity is denoted as $\alpha_j \in [0,1]$.

In the GS turbulence mitigation modeling proposed in this paper, the model is primarily based on 2D images, allowing for reasonable simplifications of the Gaussian parameters. The three-dimensional position is simplified to a two-dimensional position, keeping only the $xy$ values, assuming that all the Gaussian spheres lie on the same plane. The rotation quaternion is confined to the XY plane, meaning that the Gaussian spheres are considered to rotate only around the Z-axis. Additionally, the scaling factor $s$ is simplified to affect only the XY plane, ignoring the changes along the Z-axis.

Atmospheric turbulence degradation can be decoupled into geometric distortions (tilt) and non-isoplanatic blur (blur) in the image space. For tilt, only the mean position of the Gaussian distribution is affected. The distorted mean position $x'$ is:

$$x' = x + \Delta x \tag{5}$$

where, $x'$ represents the distorted mean position, $x$ is the true mean position, and $\Delta x$ denotes the position distortion caused by atmospheric turbulence.

For the blur effects caused by atmospheric turbulence, they can be viewed as a weighted sum of anisotropic Gaussian blur kernels $\mathbf{B}(x)$. The expression is as follows:

$$\begin{cases} \mathbf{B}(x) = \sum_{k=1}^{K} \omega_k \cdot N(x; 0, \Sigma_{\mathbf{B}_k}) \\ \Sigma_{\mathbf{B}_k} = \mathbf{R}(\theta_k) \begin{bmatrix} \sigma_{k,1}^2 & 0 \\ 0 & \sigma_{k,2}^2 \end{bmatrix} \mathbf{R}(\theta_k)^{\mathrm{T}} \end{cases} \tag{6}$$

where, $\omega_k$ is the weight of the $k$-th sum, and $\Sigma_{\mathbf{B}_k}$ is the covariance matrix of the $k$-th blur kernel. The rotation matrix corresponding to the angle $\theta_k$ is denoted as $\mathbf{R}(\theta_k)$. The covariance of the Gaussians projected onto the image plane is $\Sigma_i$. After applying the weighted blur kernels, the total covariance matrix $\Sigma_i'$ is:

$$\Sigma_i' = \Sigma_i + \sum_{k=1}^{K} \omega_k \cdot \mathbf{R}(\varphi_i) \cdot \Sigma_{\mathbf{B}_k} \cdot \mathbf{R}(\varphi_i)^{\mathrm{T}} \tag{7}$$

where, $\mathbf{R}(\varphi_i)$ represents the rotation matrix corresponding to the rotation angle $\varphi_i$ of the Gaussians.

The derivation of the above formula shows that the blur effect actually impacts the covariance matrix of the Gaussians, which in turn affects the rotation quaternion $r$ and scaling vector $s$. Therefore, this paper proposes a GS deblurring algorithm based on the non-isoplanatic blur imaging process.

The imaging process affected by the blur effect, under non-isoplanatic conditions, causes the point spread function (PSF) to vary with the spatial position $u$ in the image plane. This can be expressed by the following equation:

$$\mathbf{I}_{\mathrm{blur}}(u) = \sum_{k=1}^{K} \omega_k(u) \cdot [\mathbf{I}_{\mathrm{sharp}} \otimes \mathbf{B}(u)] \tag{8}$$

That is, for each blurred image frame, the following condition must be satisfied:

$$\begin{cases} F_{\mathrm{blur}}(\widehat{\mathbf{I}}_{\mathrm{sharp}}, k_i) \approx \mathbf{I}_i^{\mathrm{blur}} = \sum_{j=1}^{M} G_j(\Sigma_j \otimes k_i) \\ \widehat{\mathbf{I}}_{\mathrm{sharp}} \approx F_{\mathrm{deblur}}(\mathbf{I}_i^{\mathrm{sharp}}, k_i^{-1}) \end{cases} \tag{9}$$

where, $\widehat{\mathbf{I}}_{\mathrm{sharp}}$ represents the restored clear image, $G_j$ represents the $j$-th Gaussians, and $\Sigma_j \otimes k_i$ represents the interaction between the covariance of the Gaussians and the blur kernel. The corresponding variables to be optimized in the Gaussians are the rotation quaternion $r$ and the scaling vector $s$, as derived in the previous formula (3). $\mathbf{I}_i^{\mathrm{blur}}$ represents the $i$-th frame of the turbulence-degraded image, and $k_i$ represents the corresponding turbulence blur kernel of the $i$-th frame. $F_{\mathrm{blur}}$ represents the turbulence blur process based on the blur kernel. $F_{\mathrm{deblur}}$ is the deblurring process, where the inverse of the blur kernel, denoted by $k_i^{-1}$, represents the reverse blur process. During the optimization, this is achieved by implicitly optimizing the rotation quaternion $r$ and scaling vector $s$ in the Gaussians.

From the above equations, a cyclic consistency loss can be constructed:

$$L_{\mathrm{cycle}} = \frac{1}{N} \sum_{i=1}^{N} \left\| F_{\mathrm{blur}}(F_{\mathrm{deblur}}(\mathbf{I}_i^{\mathrm{blur}}, k_i^{-1}), k_i) - \mathbf{I}_i^{\mathrm{blur}} \right\| \tag{10}$$

By minimizing the cyclic consistency loss, the issue of multiple solutions in turbulence mitigation can be effectively addressed, ensuring the physical consistency between blur kernel estimation and clear image reconstruction. This provides theoretical support for end-to-end optimization in complex turbulence degradation scenarios.

*C Tilt Correction Based on Optical Flow Estimation*

In atmospheric turbulence, tilt manifest as random relative pixel displacements between each frame of the image.

Statistically, this is a zero-mean random variation process. Therefore, by calculating the optical flow between each frame, the relative displacement between images can be represented, and tilt can be corrected using the zero-mean prior. In this paper, we use RAFT to estimate the relative optical flow between images, with the detailed process illustrated in Fig. 1(b).

First, we define a base reference image as $\mathbf{I}_0$, with the other image sequence as $\{\mathbf{I}_t\}_{t=1}^{N}$. Then, RAFT is utilized to estimate the optical flow field from $\mathbf{I}_0$ to $\{\mathbf{I}_t\}_{t=1}^{N}$, as which represents the displacement of pixel $\boldsymbol{p}$ in $\mathbf{I}_0$ to its position in $\{\mathbf{I}_t\}_{t=1}^{N}$:

$$F_{t \to 0}(\boldsymbol{p}) = (\Delta x, \Delta y), \boldsymbol{p} = (x, y) \tag{11}$$

The average displacement field is obtained by calculating the mean of all the optical flow fields estimated by RAFT:

$$\bar{F}(\boldsymbol{p}) = \frac{1}{N}\sum_{t=1}^{N} F_{t \to 0}(\boldsymbol{p}) \tag{12}$$

Based on the zero-mean prior of turbulence-induced tilt, the average displacement field $\bar{F}(\boldsymbol{p})$ is the inverse of its own distortion field $F_{0 \to gt}$. Therefore, by adding the average displacement field to the distortion field, the corrected image $\mathbf{I}_{correct}$ can be obtained:

$$\mathbf{I}_{correct}(\boldsymbol{p}) = \mathbf{I}_0(\boldsymbol{p} + \bar{F}(\boldsymbol{p})) \tag{13}$$

where, $\boldsymbol{p} + \bar{F}(\boldsymbol{p})$ represents the coordinates of the sampled position in $\mathbf{I}_0$.

During the optimization process, the average displacement field $\bar{F}(\boldsymbol{p})$ is mapped to the mean position $\boldsymbol{x}$ of the Gaussians. By adjusting the position of the Gaussians, the tilt in the image can be corrected

In summary, by modeling atmospheric turbulence degradation using Gaussians, and leveraging the parameterization properties of GS, the tilt and blur in the turbulence degradation process are decoupled into different Gaussian parameters, thus achieving a more effective optimization process.

*D. Blur Kernel Weight Estimation Module*

As shown in Fig. 1(c), the Blur Kernel Estimation Module is a key component of our framework. In this module, we propose BKENet, a blur kernel weight estimation network based on the ResNeXt [9] for atmospheric turbulence deblurring. The model takes as input turbulence-blurred images and PCA blur kernel basis functions. PCA is applied to decompose the blur kernel of atmospheric turbulence into 101 basis functions, which are based on simulated blur kernels that reflect the statistical characteristics of atmospheric turbulence [5]. By adjusting the weighted combination of these basis functions, different blur kernels can be formed. BKENet is designed to estimate these spatially varying blur kernel, enabling accurate modeling and correction of the blur caused by atmospheric turbulence.

In atmospheric turbulence degradation, under small-field imaging conditions, the process can be treated as isoplanatic imaging, meaning that within this imaging range, the PSF remains approximately consistent. The isoplanatic angle $\theta$ is a key parameter for measuring the impact of atmospheric turbulence on the imaging system, defined as the maximum angle at which the wavefront distortion is approximately uniform within the field of view. The calculation formula for the isoplanatic angle [38] is:

$$\theta = 0.058 \cdot \lambda^{6/5} \cdot \left[\int_0^{\infty} C_n^2(h) \cdot h^{5/3} dh\right]^{-3/5} \tag{14}$$

where, $\lambda$ is the observation wavelength, and $C_n^2(h)$ represents the atmospheric refractive index structure constant as a function of height $h$.

Under horizontal path imaging conditions, the calculation formula for the isoplanatic angle can be simplified [38] to:

$$\theta \approx 0.53 \cdot \frac{r_0}{L} \tag{15}$$

where, $r_0$ is the Fried constant, which represents the intensity of atmospheric turbulence. The larger the value, the smoother the turbulence. $L$ is the horizontal imaging distance. Given the field of view (FOV) and image resolution, the number of pixels covered by the isoplanatic region can be calculated as:

$$\begin{cases} \alpha = \dfrac{FOV}{H \cdot W} \\ N = \dfrac{\theta_0}{\alpha} \end{cases} \tag{16}$$

where, $\alpha$ is the angular resolution of a single pixel, and $N$ is the number of pixels covered by the isoplanatic region.

Based on the calculation of the pixel size in the isoplanatic region, the image can be divided into several local isoplanatic regions, where the blur kernel within each region can be assumed to be consistent. This significantly reduces the number of parameters that need to be optimized and enhances the stability of the optimization process.

A key innovation in this work is leveraging isoplanatic regions to simplify blur kernel estimation. By focusing on isoplanatic regions, the model can estimate a smaller set of parameters, reducing the need for complex spatial variations in the blur kernel. For a 512×512 image, this leads to a significant reduction in the number of estimated parameters—from 100×512×512 to 100×32×32, where 100 represents the basis functions and 32×32 corresponds to the resolution of the isoplanatic regions. This reduction in parameter size decreases computational complexity and enhances optimization efficiency, enabling faster and more stable results without compromising restoration quality.

In addition, BKENet incorporates two further innovations to improve the blur kernel estimation: weight positivity constraints and sparsity regularization. The weight positivity constraint ensures that the blur kernel weights remain non-negative, which is essential for physically meaningful kernel estimation. Meanwhile, sparsity regularization encourages the model to focus on a small set of significant basis functions, reducing the number of active parameters and improving efficiency. This regularization also helps prevent overfitting, allowing the model to learn more compact and interpretable representations of the blur kernel. The influence of positive weight constraints, and sparsity regularization are shown in Fig. 5(d)

By integrating isoplanatic region-based parameter simplification, positive weight constraints, and sparsity regularization, BKENet enhances its robustness and efficiency, enabling precise estimation of blur kernel weights with minimized computational complexity.

The process of GSTurb framework is summarized in Algorithm 1.

---

**Algorithm 1: GSTurb**

**Input**:
Degraded image $\mathbf{I}_{degrade}$
**Output**:
Restored image $\mathbf{I}_{restored}$

**1. Initialize Gaussian Splatting (GS) Parameters**
　　GS parameters←initialize_GS_parameters()

**2. Tilt Correction Module**
**Input:** Degraded image $\mathbf{I}_{degraded}$
**Output:** Tilt-corrected image $\mathbf{I}_{tilt\_corrected}$
　　$\mathbf{I}_{ref}$←select_reference_image ($\mathbf{I}_{degraded}$)
　　optical_flow←estimate_optical_flow ($\mathbf{I}_{ref}$, $\mathbf{I}_{degraded}$)
　　averaged_flow←average_optical_flow (optical_flow)
　　$\mathbf{I}_{tilt\_corrected}$←correct_tilt ($\mathbf{I}_{ref}$, averaged_flow)
**Return** $\mathbf{I}_{tilt\_corrected}$

**3. Blur Kernel Estimation Module**
**Input:** Tilt-corrected image $\mathbf{I}_{tilt\_corrected}$
**Output:** Blur Kernel $k_i$
　　$k_i$←BKENet ($\mathbf{I}_{tilt\_corrected}$)
**Return** $k_i$

**4. Optimization Module**
**Input:** Tilt-corrected image $\mathbf{I}_{tilt\_corrected}$, Blur Kernel $k_i$
　　converged←False
**While not converge:**
　　GS parameters←optimize_GS_parameters ($\mathbf{I}_{tilt\_corrected}$)
　　$\mathbf{I}_{restored}$←apply_GS ($\mathbf{I}_{tilt\_corrected}$, GS parameters)
　　loss←calculate_cyclic_consistency_loss ($\mathbf{I}_{restored}$, $\mathbf{I}_{tilt\_corrected}$)
**If loss<threshold:**
　　converged←True
**End While**
**Return** $\mathbf{I}_{restored}$

---

## IV. EXPERIMENTS

In this section, comprehensive experiments are conducted to evaluate the performance of GSTurb and other methods.

### A. Datasets

**Synthetic Datasets**: due to the difficulty of obtaining large-scale real images corresponding to atmospheric turbulence-degraded images under outdoor conditions, the atmospheric turbulence degradation image synthesis dataset ATSyn [28] was released in 2025, which has shown good performance in model training. Therefore, in this study, the ATSyn-static dataset was used, which includes ground truth images, images with only tilt, images with only blur, and images with both distortion and blur fusion. This allows for a comprehensive evaluation of the degradation mitigation algorithm's performance. In the training process of this paper, only the blurred images were used as training data. The dataset is divided into a training set and a test set. The training set contains 1,636 categories, with 50 different blurred images in each category, totaling 81,800 images. The test set contains 1,000 categories, with 163 randomly selected categories, totaling 8,150 images used as the test set during training.

**Real-World Datasets**: images degraded by real atmospheric turbulence can be used to test the performance and generalization of the proposed model in real-world environments. This study uses two real-world datasets: TSRWGAN Real-World Datasets [31] and CLEAR [32]. The TSRWGAN Real-World Datasets contains 6,316 videos with a total of 606,336 frames of images, captured in real scenes in Australia. The CLEAR dataset includes 8 static scenes, with pseudo-ground-truth data generated through complex wavelet image fusion. Overall, these datasets cover a wide range of real turbulence conditions.

### B. Optical-Flow Tilt Correction Experiment

As defined by Equation (16), the tilt in the reference image can be corrected by computing the mean of the relative optical flow fields. The correction results are shown in Fig. 2. Fig. 2a displays the relative optical flow fields between the reference image and the other images, with frames 1, 2, ..., 49 representing the relative optical flow fields of each image compared to the base reference image $I_0$. After averaging these relative optical flow fields, the result is shown in Fig. 2a-(b). This result is the inverse of the ground-truth optical flow field Fig. 2a-(a) of $I_0$, and when this result is added to the ground-truth optical flow field, we can see that the ground-truth optical flow field of $I_0$ is corrected, as shown in Fig. 2a-(c), where the optical flow field are close to zero. This experimental result proves that multi-frame optical flow field summation can correct the tilt in the base reference image $I_0$.

In real atmospheric turbulence degraded images, tilt and image blur are coupled. Therefore, it is necessary to verify whether the blur effect influences the optical flow field caused by the tilt. The experimental results are shown in Fig. 2c. In this figure, (a) and (b) represent the total summation of multi-frame relative optical flow fields under the tilt-blur coupling and tilt-only conditions, respectively, while (c) and (d) represent the corresponding single-frame relative optical flow fields. From the comparison of (c) and (d), it can be observed that the presence of blur effect has little impact on the overall changes in the optical flow field. However, for some local regions, the blur effect causes the optical flow field to show a more regular and smoother variation trend, indicating a local influence of the blur effect on the optical flow field computation. From the comparison of (a) and (b), it can be seen that in the case of summing multi-frame relative optical flow fields, the impact of the blur effect is almost negligible, and the two results are nearly identical. This shows that the process of summing multi-frame relative optical flow fields weakens the influence of the blur effect on the optical flow field computation. Through the above experiments, it has been demonstrated that the method of calculating the optical flow field using RAFT can effectively correct the tilt in the images. The correction results are shown in Fig. 2b. In this figure, (a) is the ground truth image, (b) shows the residual between the tilt image and the ground truth image before correction, and (c) shows the residual after correction. The PSNR of the image improved from 30.35 dB to 37.73 dB, a 24.31% improvement.

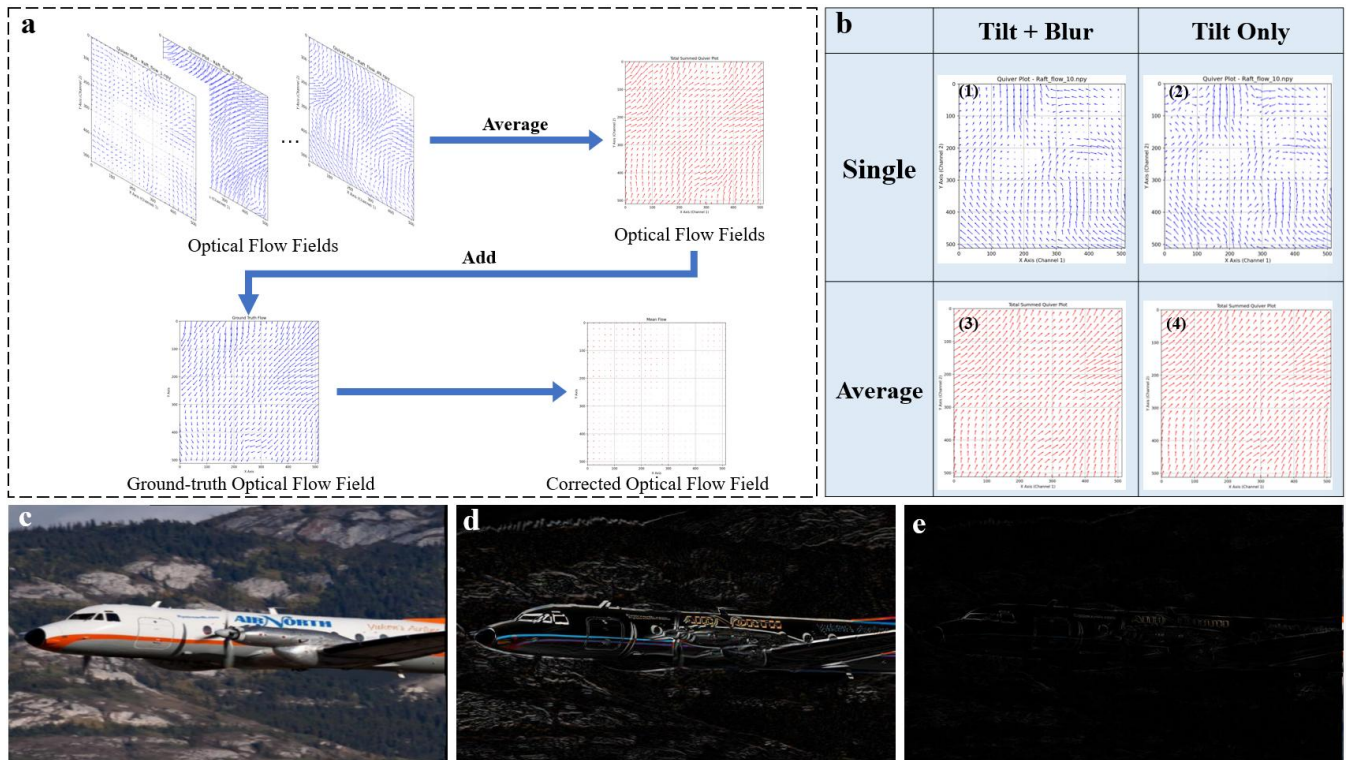

Fig. 2. Tilt correction experiment results. Fig. 2a illustrates the process and experimental results of multi-frame optical-flow tilt correction. Fig. 2b illustrates the impact of the blur effect on the optical flow field estimation of individual frames and the average optical flow fields. Fig. 2c to 2e shows the image residuals before and after optical flow field correction, where c is the undistorted image, d shows the residual before correction, and e shows the residual after correction.

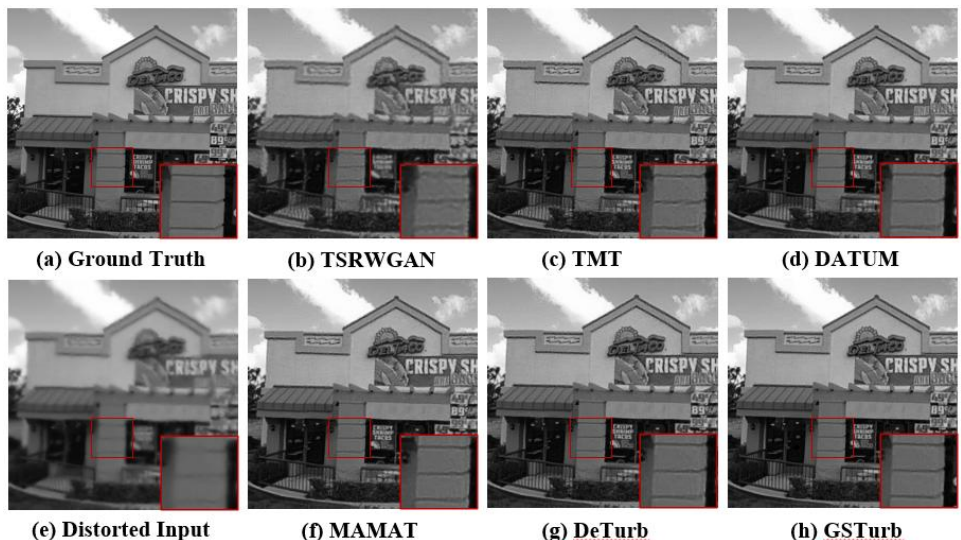

Fig. 3. Comparison of Mitigation Results on ATSyn-statics Datasets. From (a) to (h): (a) and (e) show the ground truth and the original degraded image, respectively. The mitigation results of (b) TSRWGAN, (c) TMT, (d) DATUM, (e) Distorted Input, (f) MAMAT, (g) DeTurb, (h) our GSTurb, respectively. Our GSTurb can produce the sharpest image.

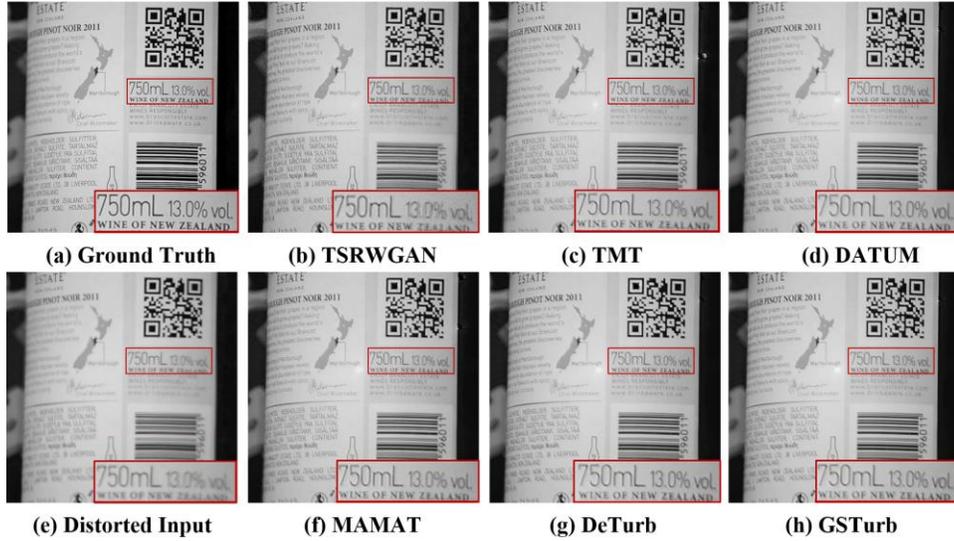

Fig. 4. Comparison of Mitigation Results on CLEAR Datasets. From (a) to (h): (a) and (e) show the ground truth and the original degraded image, respectively. The mitigation results of (b) TSRWGAN, (c) TMT, (d) DATUM, (e) Distorted Input, (f) MAMAT, (g) DeTurb, (h) our GSTurb, respectively. Our GSTurb can produce the sharpest image.

*C. BKENet Training*

To train the proposed model, the SGD optimizer and a cosine annealing learning rate scheduler [34] are used. The initial learning rate was set to 0.1, with a batch size of 14. The number of training iterations was 1.2 million. The model training was distributed across 4 RTX 4090 GPUs, using ground truth and blurred image pairs for training. The training and validation sets consisted of a total of 80,000 images, divided into a 9:1 ratio, with random shuffling.

We evaluated and compared our method with five other models specifically designed for atmospheric turbulence mitigation: TSRWGAN [31], TMT [27], DATUM [28], MAMAT [35], and DeTurb [36]. These models were trained following the original papers and publicly available code.

*D. Synthetic Dataset Results*

For the synthetic dataset ATSyn-statics, which includes ground truth, the mitigation results were evaluated using PSNR and SSIM. Table I presents the average calculation results, computed by averaging all the results in the test set. From the data, we observe that the traditional Generative Adversarial Network (GAN) model TSRWGAN achieved a PSNR of only 23.16 dB and SSIM of 0.7016. With slightly more complex structural models like TMT, the PSNR increased to 24.01 dB and SSIM to 0.7674. The three-step restoration models, DATUM, MAMAT, and DeTurb, respectively, improved PSNR to 26.15 dB, 26.21 dB, and 26.37 dB, and SSIM to 0.8244, 0.8325, and 0.8256. The GSTurb method proposed in this paper, which combines optical flow correction for tilt and Gaussian splatting for modeling non-uniform blur, significantly improved performance, achieving a PSNR of 27.67 dB and SSIM of 0.8735. Compared to DeTurb, the PSNR improved by approximately 1.3 dB (a 4.5% increase), and the SSIM increased by nearly 0.048 (a 5.8% improvement). Compared to the weakest TSRWGAN, the improvement was more noticeable, with PSNR increasing by 4.51 dB (a 19.5% increase) and SSIM rising by 0.172 (a 24.5% increase). This improvement validates that combining optical flow-guided tilt correction with Gaussian splatting for blur modeling effectively corrects the random geometric shifts caused by atmospheric turbulence and accurately estimates the spatially varying PSF, thereby significantly improving the reconstruction quality of the synthetic dataset. The mitigation quality of each method is shown in Fig. 3.

TABLE I
QUANTITATIVE COMPARISONS (AVERAGE PSNR/SSIM) WITH SOTA ON ATSYN-STATIC

| Benchmark | ATSyn-static | |
|---|---|---|
| Methods | PSNR | SSIM |
| TSRWGAN | 23.16 | 0.7016 |
| TMT | 24.01 | 0.7674 |
| DATUM | 26.15 | 0.8244 |
| MAMAT | 26.21 | 0.8325 |
| DeTurb | 26.37 | 0.8256 |
| GSTurb(ours) | **27.67** | **0.8735** |

*E. Real-World Datasets Results*

For the real dataset, the CLEAR dataset includes pseudo-ground truth, so PSNR and SSIM can be used as evaluation metrics. The TSRWGAN Real-World Datasets are atmospheric turbulence-degraded videos captured from real-world scenes, lacking ground truth for comparison. In this paper, BRISQUE [37] and GCL scores are used as no-reference evaluation metrics. The BRISQUE metric (Blind/Referenceless Image Spatial Quality Evaluator) is a no-reference image quality assessment metric based on the spatial structural features of an image. It evaluates quality by analyzing the natural scene statistics in the image. A higher BRISQUE score indicates poorer image quality. The GCL metric (Gradient Magnitude Contrast, GCL) is a no-reference indicator that evaluates image contrast and sharpness. GCL primarily measures image details and clarity by calculating the gradient magnitude of the image. It reflects the preservation of image details, particularly high-frequency information, by analyzing edge variations. A higher GCL score indicates better image clarity and more detailed content.

TABLE II
QUANTITATIVE COMPARISONS (AVERAGE PSNR/SSIM) WITH SOTA ON CLEAR DATASETS

| Turbulence Level | Weak | | Medium | | Strong | | Overall | |
|---|---|---|---|---|---|---|---|---|
| Methods | PSNR | SSIM | PSNR | SSIM | PSNR | SSIM | PSNR | SSIM |
| TSRWGAN | 23.38 | 0.8332 | 20.56 | 0.6993 | 18.07 | 0.6523 | 20.67 | 0.7282 |
| TMT | 23.53 | 0.8355 | 21.86 | 0.7535 | 20.32 | 0.7158 | 21.90 | 0.7682 |
| DATUM | 25.10 | 0.8492 | 23.05 | 0.7981 | 21.65 | 0.7404 | 23.26 | 0.7959 |
| MAMAT | 25.63 | 0.8512 | 23.37 | 0.8023 | 21.86 | 0.7327 | 23.62 | 0.7954 |
| DeTurb | 25.89 | 0.8324 | 23.85 | 0.8037 | 21.05 | 0.7519 | 23.53 | 0.7960 |
| GSTurb(ours) | **26.33** | **0.8536** | **24.57** | **0.8358** | **23.35** | **0.8073** | **24.75** | **0.8322** |

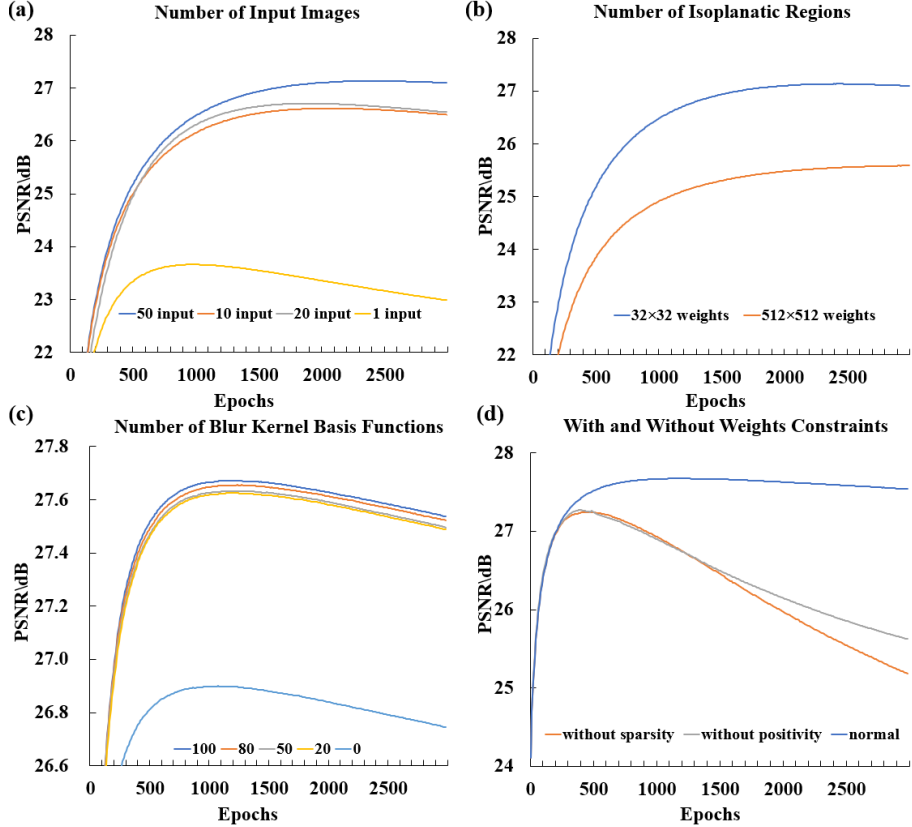

Fig. 5. Performance Evaluation of GSTurb with Different Input Configurations. (a) PSNR comparison of blurred image restoration for varying input image quantities (1, 10, 20, and 50). (b) PSNR comparison for different numbers of isoplanatic regions (32×32 and 512×512). (c) PSNR comparison for varying numbers of blur kernel basis functions. (d) PSNR comparison for different weights constraints.

The CLEAR datasets are categorized into three levels based on turbulence intensity: Weak, Medium, and Strong, and the overall average performance is summarized in Table II. The results show that as turbulence intensity increases, the PSNR/SSIM of all methods decreases, but our method consistently outperforms the others. Under weak turbulence conditions, GSTurb achieves a PSNR/SSIM of 26.33 dB/0.8536, outperforming DeTurb by 0.44 dB (a 1.7% increase) and 0.0212 (a 2.5% increase); under medium turbulence, it reaches 24.57 dB/0.8358, 0.72 dB (a 3.0% increase) and 0.0321 (a 4.0% increase) higher than DeTurb; under strong turbulence, it achieves 23.35 dB/0.8073, surpassing DeTurb by 1.30 dB (a 5.9% improvement) and 0.0518 (a 6.9% improvement). The overall average PSNR is 24.75 dB and SSIM is 0.8422, outperforming DeTurb, by 1.22 dB (a 5.2% increase) and 0.0462 (a 5.8% improvement), respectively. These results demonstrate that the proposed model can consistently estimate spatially varying blur kernels in complex real-world turbulent environments, and through iterative optimization, it achieves clearer results. The performance is particularly notable under strong turbulence, suggesting that our method is more robust to random distortions and blur coupling. The mitigation quality of each method is show in Fig. 4.

Table III shows the comparison of results on TSRWGAN Real-World Datasets. The results show that TSRWGAN has a BRISQUE of 44.86 and a GCL of 12.10. DeTurb reduces BRISQUE to 39.50 and increases GCL to 16.98. GSTurb further improves the results, reducing BRISQUE to 39.13 and increasing GCL to 17.10, achieving the lowest BRISQUE and highest GCL among all algorithms. Compared to DeTurb, BRISQUE decreases by 0.37, and GCL increases by 0.12. Compared to TSRWGAN, BRISQUE decreases by 5.73, and GCL increases by 5.00, demonstrating that image naturalness and local contrast can be improved even under real, no-reference conditions.

TABLE III
QUANTITATIVE COMPARISONS (AVERAGE BRISQUE/GCL) WITH SOTA ON TSRWGAN REAL-WORLD DATASETS

| Benchmark | TSRWGAN Real-World Datasets | |
|---|---|---|
| Methods | BRISQUE | GCL |
| TSRWGAN | 44.86 | 12.10 |
| TMT | 42.15 | 13.95 |
| DATUM | 40.43 | 15.78 |
| MAMAT | 40.32 | 16.32 |
| DeTurb | 39.50 | 16.98 |
| **GSTurb(ours)** | **39.13** | **17.10** |

*F. Ablation Study*

The number of input images and the division of isoplanatic regions significantly impact the mitigation results. Additionally, the number of blur kernel basis functions plays a crucial role in image restoration performance. The positive weight and sparsity constraints on the blur kernel basis functions also have a significant impact on the restoration quality. Fig. 5 compares the PSNR of the blurred image restoration under different conditions. In (a), the PSNR of blurred image restoration is compared for input image quantities of 1, 10, 20, and 50. In (b), the PSNR comparison is made for different numbers of isoplanatic region of 32×32 and 512×512. In (c), the PSNR comparison is shown for varying numbers of blur kernel basis functions. In (d), the PSNR comparison is shown for different weight constraints.

From Fig. 5(a), it can be seen that as the number of input images increases, the PSNR of blurred image restoration gradually improves, with the maximum PSNR reaching 23.66 dB, 26.61 dB, 26.70 dB, and 27.67 dB, respectively. Additionally, the optimization curve becomes smoother, and the degree of image degradation in the later stages of optimization is smaller. This improvement is due to the randomness and inhomogeneity of atmospheric turbulence degradation, where the blur degree in the same local region of an image varies under different blur conditions. As the number of input images increases, more information is for image restoration, resulting in better mitigation performance. From Fig. 5(b), it can be observed that when the number of isoplanatic regions is 32 × 32, or 16 pixels ×16 pixels per region, the maximum PSNR of the blurred image restoration reaches 27.67 dB, which is higher than the 25.59 dB achieved when the number of isoplanatic regions is 512×512. Dividing the isoplanatic regions significantly reduces the number of parameters required to optimize the blur kernel basis function weights, making the optimization process more stable without causing significant information loss.

From Fig. 5(c), it can be observed that as the number of blur kernel basis functions increases, the PSNR of blurred image restoration improves. The maximum PSNR values gradually increase, from 26.90 to 27.67, with higher numbers of basis functions leading to a more precise restoration process. When the number of basis functions is set to 0, a fixed blur kernel is used. The data results show that using a linear combination of basis functions leads to better restoration performance than using a single fixed blur kernel. This improvement is attributed to the increased model capacity for capturing more detailed blur characteristics, allowing for better reconstruction of the blurred image.

From Fig. 5(d), the results clearly demonstrate the impact of positivity and sparsity constraints on the performance of the model. The normal curve, which represents the case with both positivity and sparsity constraints, shows the best performance with a higher and more stable PSNR over epochs. In contrast, the without sparsity and without positivity curves reveal slower convergence and lower final PSNR values, indicating that the absence of these constraints negatively affects the model's restoration quality. These findings highlight that both constraints are essential for improving the model's efficiency and stability, significantly enhancing its ability to estimate blur kernel weights accurately and restore images with better quality.

## IV. Conclsion

This paper proposes a framework for atmospheric turbulence mitigation based on Gaussian Splatting (GSTurb). By combining optical flow correction and deep blur kernel estimation, we successfully improve the restoration of tilt and blur under joint optimization of multiple frames. Experimental results demonstrate that the proposed GSTurb method achieves performance improvements across multiple datasets, particularly in commonly used image quality assessment metrics such as PSNR and SSIM. By introducing GS-based turbulence modeling and non-rigid optical flow correction, we can effectively handle complex spatially varying blurs and further optimize the recovery process under multi-frame input. Furthermore, the accurate estimation of blur kernel basis function weights significantly enhances the clarity of the restored images, proving the robustness of our method.

However, this study is limited to static scene mitigation, and while BKENet shows good generalization ability, it is still constrained by the dataset used for training. Future work will focus on extending the proposed framework to dynamic scene mitigation and incorporating atmospheric turbulence wavefront detection. These improvements aim to further enhance the generalization and practical applicability of the method.

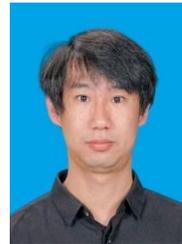

**Hanliang Du** received the M.S. degree from Xi'an Jiaotong University, Xi'an, China. He is currently pursuing the Ph.D. degree with the University of Macau, Macau, China. His research interests include three-dimensional (3D) Gaussian splatting, atmospheric turbulence mitigation, and computational imaging. His current work focuses on physics-guided learning frameworks for turbulence-degraded image restoration and dynamic scene reconstruction.

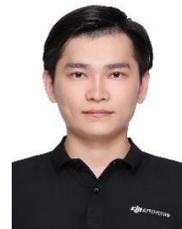

**Zhangji Lu** received the Ph.D. degree in electronic and computer engineering from the Hong Kong University of Science and Technology, Hong Kong, China, in 2021. He was with DJI Innovations from 2021 to 2024 as a Perception Algorithm Engineer, where he worked on autonomous driving and visual perception systems. Since November 2024, he has been with the College of Physics and Optoelectronic Engineering, Shenzhen University, China. His current research interests include simultaneous localization and mapping (SLAM), multi-view stereo (MVS), fringe projection profilometry (FFP), 3-D Gaussian splatting, and industrial anomaly detection.

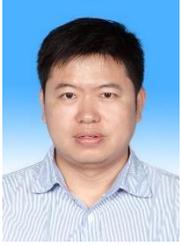

**Xiaoli Liu** received the Ph.D. degree in precision testing and measurement technology and instruments from Tianjin University, Tianjin, China, in 2008. He is currently a Professor and Doctoral Supervisor with the College of Physics and Optoelectronic Engineering, Shenzhen University, China. He is a Topic Editor of *Acta Optica Sinica*, a Member of the Visual Inspection Committee of the Chinese Society of Image and Graphics, a Director of the Graphical and Image Branch of the China Instrument and Control Society, and a Member of the SPIE. His current research interests include three-dimensional optical digital imaging and modeling, computational optical field 3-D imaging, laser Doppler displacement measurement, white-light interferometric metrology, structured-light 3-D measurement, and machine vision–based optical inspection.

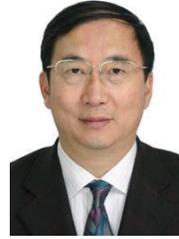

**Qifeng Yu** (Fellow, Chinese Academy of Sciences) received the Ph.D. degree in applied optics from the Bremen Institute for Applied Beam Technology (BIAS), University of Bremen, Germany, in 1995. His research interests include experimental mechanics, optical measurement, and image-based deformation and motion analysis. He has developed fundamental theories and technologies for large-scale deformation and motion measurement, which have been successfully applied to manned spaceflight and national defense. Prof. Yu is an Academician of the Chinese Academy of Sciences and a recipient of the National Technological Invention Award (Second Prize) and the Military Science and Technology Progress Award (Second Prize)

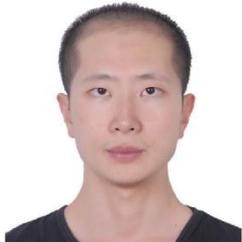

**Zewei Cai** received Ph.D. in Optical Engineering from Shenzhen University in 2017, focusing on research in computational light field imaging and measurement. His work explores the mechanisms and methods of light field modulation, involving light field inversion models such as scattering encoding, aperture modulation, and intensity transmission. He has also developed structured light field 3D measurement technologies and equipment by integrating structured illumination control, achieving high-dimensional light signal joint transmission and demodulation. These technologies are applied in optical measurement, lensless imaging, microscopic imaging, and coherent measurement.

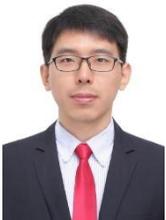

**Qijian Tang** received the Ph.D. degrees in precision instrumentation and optoelectronic engineering from Tianjin University, Tianjin, China, in 2015, respectively. From 2013 to 2014, he was a joint Ph.D. student with Brunel University London, U.K. He is currently an Associate Professor and Ph.D. Supervisor with the College of Physics and Optoelectronic Engineering, Shenzhen University, Shenzhen, China. His research interests include optical three-dimensional (3-D) imaging and measurement, Fourier ptychographic imaging, and computational optical imaging and metrology.